\documentclass[10pt,twocolumn,letterpaper]{article}

\usepackage{cvpr}
\usepackage{times}
\usepackage{epsfig}
\usepackage{graphicx}
\usepackage{amsmath}
\usepackage{amssymb}
\usepackage{bm}
\usepackage[english]{babel}

\usepackage[pagebackref=true,breaklinks=true,letterpaper=true,colorlinks,bookmarks=false]{hyperref}

\cvprfinalcopy 

\def\cvprPaperID{2280} 

\newcommand{\norm}[1]{\left\lVert#1\right\rVert}
\newcommand{\matr}[1]{\bm{#1}}
\newcommand{\argmin}{\operatornamewithlimits{argmin}}

\ifcvprfinal\fi
\begin{document}

\title{Hierarchical Piecewise-Constant Super-regions}

\author{Imanol Luengo and Andrew P. French\\
School of Computer Science\\
University of Nottingham\\
Nottingham, UK, NG8 1BB\\
{\tt\small \parbox{3.1cm}{\{imanol.luengo,\\ andrew.p.french\}}@nottingham.ac.uk}
\and
Mark Basham\\
Diamond Light Source Ltd\\
Harwell Science \& Innovation Campus\\
Didcot, UK, OX11 0DE\\
{\tt\small mark.basham@diamond.ac.uk}
}

\maketitle

\providecommand{\keywords}[1]{\textbf{\textit{Keywords---}} #1}

\begin{abstract}
   Recent applications in computer vision have come to heavily rely on superpixel over-segmentation as a pre-processing step for higher level vision tasks, such as object recognition, image labelling or image segmentation. Here we present a new superpixel algorithm called Hierarchical Piecewise-Constant Super-regions (HPCS), which not only obtains superpixels comparable to the state-of-the-art, but can also be applied hierarchically to form what we call \textit{n}-th order super-regions. In essence, a Markov Random Field (MRF)-based anisotropic denoising formulation over the quantized feature space is adopted to form piecewise-constant image regions, which are then combined with a graph-based split \& merge post-processing step to form superpixels. The graph and quantized feature based formulation of the problem allows us to generalize it hierarchically to preserve boundary adherence with fewer superpixels. Experimental results show that, despite the simplicity of our framework, it is able to provide high quality superpixels, and to hierarchically apply them to form layers of over-segmentation, each with a decreasing number of superpixels, while maintaining the same desired properties (such as adherence to strong image edges). The algorithm is also memory efficient and has a low computational cost.
\end{abstract}

\keywords{superpixels, super-regions, hierarchy, segmentation, image}

\section{Introduction}

There is an increasing trend to use superpixels as building blocks for many computer vision applications such as image segmentation \cite{ren2003learning}, image parsing \cite{tighe2010superparsing} or semantic labelling \cite{kohli2009robust} and object tracking \cite{wang2011superpixel}. Superpixel algorithms group pixels into perceptually meaningful regions, which are more aligned with the human visual cognition system. They not only reduce the redundancy and  noise effects in the standard individual pixel grid, but also are especially useful for high computational cost problems, as operating in a superpixel graph reduces dimensionality of the problem (and thus the computational complexity) by several orders of magnitude with respect to the full pixel grid. To enable this reduction in resolution to be helpful, there are some understood properties that a superpixel algorithm should offer in order provide quality end results in the subsequent higher-level applications:
\begin{enumerate}
\item Superpixels should adhere to image boundaries.
\item Each superpixel should be contained in a unique higher level object. This is, a superpixel should not overlap more than one object in the image.
\item In most applications superpixels are used as a preprocessing step; therefore they should be fast to compute and memory efficient.
\end{enumerate}

\begin{figure}[t]
\begin{center}
   \includegraphics[width=\linewidth]{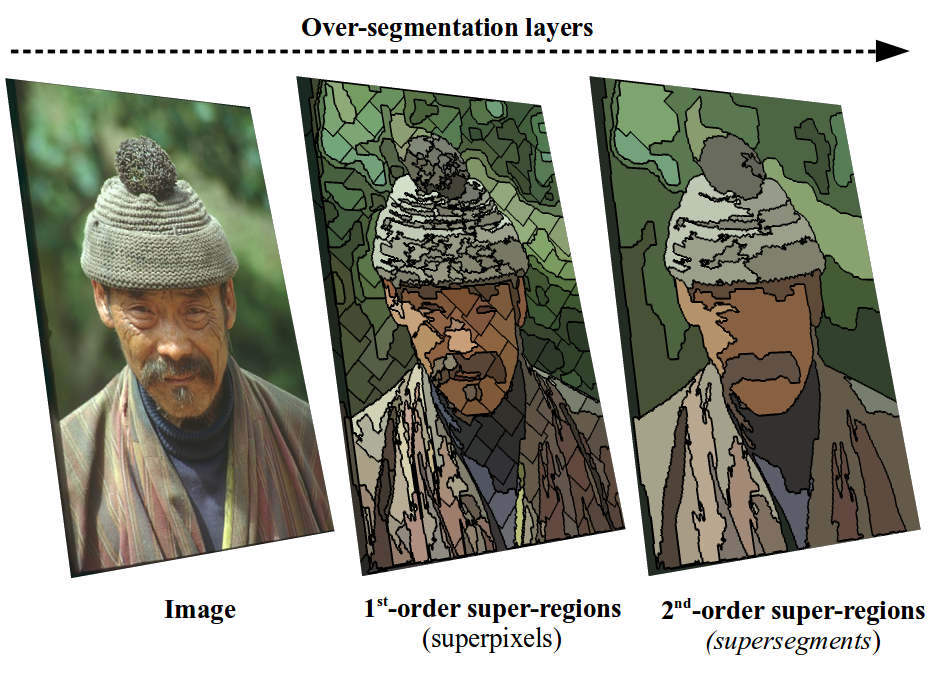}
\end{center}
   \caption{Overview of the \textit{Hierarchical Piecewise-Constant Super-regions (HPCS)}. The output of each layer is used as an input for the next one, yielding increasingly larger regions while preserving image's strong edges.}
\label{fig:abstract}
\end{figure}

The existing superpixel algorithms \cite{achanta2012slic}\cite{liu2011entropy}\cite{van2012seeds}\cite{Li2015CVPR}\cite{levinshtein2009turbopixels} efficiently meet the needs of the different computer vision problems. The more superpixels that are extracted from an image, the higher quality they get, and thus, the better the results of the subsequent higher level applications become. It is then desired to find a balance between the quality and the dimensionality of the image representation. However, for high-dimensional images such as High Definition 4K images or 3D biomedical volumes, the amount of superpixels needed to maintain the boundary adherence increases rapidly with the size of the dataset. This introduces a new problem: while a small amount of large, traditional superpixels wouldn't have enough boundary adherence to ensure good results in later processing, a large number of small superpixels would start to lose their interesting perceptual characteristics as they would describe only a small local region, and there may be too many of them to be of practical use.

To address this problem, we present a hierarchical over-segmentation framework, \textit{Hierarchical Piecewise-Constant ``Super-regions'' (HPCS)}, that allows us to generalize the superpixel over-segmentation as a hierarchical process where each layer of the hierarchy outputs a decreasing amount of superpixels, while maintaining the desired quality (such as boundary adherence). We name the hierarchical process as a \textit{n}th-order super-region hierarchy, with the original image being the $0$th-order super-region and the standard superpixel segmentation the $1$st-order super-region. Further orders (layers) in the hierarchy tend to produce larger regions that maintain the strong boundaries from the previous layer. Figure \ref{fig:abstract} overviews this process.


\subsection{Contributions}

Our work has two main contributions:

\begin{enumerate}
\item A new superpixel over-segmentation algorithm, formulated as a global anisotropic denoising-based energy minimization framework. Our algorithm, despite its simple form, generally performs as well or better than most state-of-the art algorithms, and is quick to compute and memory efficient.

\item A new hierarchical over-segmentation generalization that allows us to create hierarchical layers of over-segmentations with decreasing number of superpixels while maintaining desired superpixel properties.
\end{enumerate}

We qualitatively and quantitatively demonstrate the empirical validity of our algorithm, both to create state-of-the-art superpixels and to reduce the number of superpixels needed to describe an image. We also show the validity of our hierarchical framework as a post-processing step to reduce the number of superpixels of other over-segmentation methods. And last, but not least, we discuss the applicability of this hierarchical formulation in the trending inclusion of Higher Order potentials for MRF problems \cite{kohli2007p3}\cite{kohli2009robust}, such as the recent Associative Hierarchical Random Fields \cite{russell2009associative}, which is of particular interest.


\section{Related work}

As \cite{achanta2012slic} and \cite{van2012seeds} we split the previous existing algorithms into three different categories: superpixels from a graph formulation by gradually adding cuts, superpixels grown from initialized centers, and superpixels extracted by moving a predefined set of boundaries.

Graph-based methods represent the image as a graph of pixels in a 4 or 8-neighbouring system and calculate similarities between adjacent pixels. For example, Normalized Cuts \cite{shi2000normalized} globally minimizes an objective function by recursively finding the optimal partition in the normalized Laplacian graph. While giving good results, the algorithm is computationally expensive. An alternative approach is an agglomerative clustering algorithm from Felzenszwalb and Huttenlocher \cite{felzenszwalb2004efficient} which is faster than Normalized Cuts. However, it can produce superpixels with very irregular shapes and sizes which is not always desirable. Moore \etal introduced Superpixel Lattices (SL) \cite{moore2008superpixel} that find optimal horizontal and vertical paths in a graph from a boundary map. Recently, Topology Preserving Regular superpixels (TPR) \cite{tang2012topology} improve SL by finding shortest paths. SL and TPR, however, both depend on a precomputed boundary map and the quality of it directly reflects their performance. Veksler and Boykov \cite{veksler2010superpixels} managed to generate superpixels by placing overlapped patches over the image and assigning each pixel to one of them. They formulate the problem in a MRF framework whose solution is inferred with Graph Cuts \cite{boykov2004experimental}. In 2011, Liu \etal \cite{liu2011entropy} introduced Entropy Rate superpixels (ERS), a graph-based clustering method of the entropy rate of a random walk, balanced by an energy that encourages superpixels of similar size. ERS superpixels are one of the most powerful, and they are able to detect boundaries that other superpixels tend to smooth.

Region growing methods start by using a predefined set of seed points to grow superpixels using different techniques. Perhaps a classic example of this is watershed segmentation \cite{vincent1991watersheds}. Using the gradient image, superpixels are created by flooding from the seed points in the gradient plane. An alternative approach would be QuickShift \cite{vedaldi2008quick}, which is itself a fast approximation of MeanShift \cite{comaniciu2002mean} and are both mode seeking algorithms. While their results achieve good boundary adherence, they are quite computationally expensive. Another seed-based approach is Turpopixels (TP) \cite{levinshtein2009turbopixels} which grows geometric flows from seeds until superpixels are created. Recently introduced by Achanta \etal is Simple Linear Iterative  Clustering (SLIC) \cite{achanta2012slic}, perhaps one of the most widely known superpixel algorithm due to its simple yet powerful formulation, which performs a fast variation of \textit{k}-means clustering in superpixel windows. Another recent introduction is the SEEDS algorithm \cite{van2012seeds}, which extracts superpixels by moving a predefined set of pixel boundaries in an energy maximization framework that encourages color homogeneity and shape regularity. Last, a new introduction from last year, Linear Spectral Clustering (LSC) \cite{Li2015CVPR} has been proven extremely powerful and yet efficient by exceeding most of the state-of-the-art algorithm results in common benchmarks by adopting the normalized cuts formulation and approximating the similarity metric by a kernel function, leading to an explicit mapping of the pixels into a high dimensional feature space. The most related superpixel algorithm to the work in this paper is that of Veksler and Boykov \cite{veksler2010superpixels}, as they also formulate the superpixel over-segmentation approach in a MRF framework. Their formulation, however, relies in sampled patches and uses only gray-scale information (to make it efficient). Our algorithm, as shown in section \ref{sec:experiments}, produces less compact superpixels, but achieves much better boundary adherence while being more efficient.

Superpixel algorithms are usually formulated as constrained frameworks where the number of superpixels $N$ plays an important role in the final output. As seen above, this constraint is introduced into the problem by different approaches such as initializing a grid of $N$ uniform superpixels, $N$ seed points, $N$ uniform patches or as stopping criterion when the solution reaches $N$ connected regions. Here however, we will formulate the image over-segmentation problem as an unconstrained optimization algorithm (in the number of superpixels), where the number of superpixels is later enforced as a post-processing split \& merge step. This allows us to provide a more general framework with applications to other computer vision problems such as interactive ND-image segmentation or hierarchical semantic labelling by exploiting their inherent hierarchical nature.


\section{HPCSuper-regions}

In this section we will present our super-region segmentation algorithm, which not only produces high quality superpixels, but can also be generalized as a powerful hierarchical over-segmentation framework. The HPCS algorithm is based on the widely studied anisotropic denoising methods \cite{beck2009fast}, which are an essential pre-processing step in many computer vision applications, and provide piecewise-smooth images preserving strong edges. By combining anisotropic denoising methods with the current belief that a few quantized features can encode enough discriminative information to classify whole datasets (widely used in bag-of-word feature models, for example), we will formulate the image over-segmentation as a graph-based piecewise-constant denoising in the quantized feature space and solve it in an energy minimization framework. The following sections will be structured as follows: section \ref{sec:superpixel} reviews the first layer of the over-segmentation framework, applied to extract superpixels from a given image, then section \ref{sec:superregion} will generalize the framework to further layers in the hierarchy.

\subsection{Preliminaries}

\textbf{Markov Random Fields (MRF)} have been widely applied in computer vision problems, as many of them can be stated as labelling problems. Given an undirected graph $\mathcal{G} = (\mathcal{V}, \mathcal{E})$, where $\mathcal{V}$ are the vertex (or nodes) and $\mathcal{E}$ is the edge set, and a finite set of labels $\mathcal{L}$, the task is to assign the optimal label $l \in \mathcal{L}$ to each $v \in \mathcal{V}$. The general form of a 2nd order MRF enforces unary $\psi_p$ and pairwise $\psi_{pq}$ constrains to the set of nodes and edges,
\begin{equation}\label{eq:mrf}
E(\matr{l}) = \sum_{p\in\mathcal{V}} \psi_p(l_p) + \lambda \sum_{p,q\in\mathcal{E}} w_{pq} \cdot \psi_{pq}(l_p, l_q)
\end{equation}
where $w_{pq}$ is a weighting coefficient, $\psi_p(l_p)$ express how likely a node $p$ is to be labelled as $l_p$ and $\psi_{pq}(lp, lq)$ express how likely two neighbouring nodes $p$ and $q$ are to be labelled as $l_p$ and $l_q$. Minimizing $E$ yields the optimal labelling $\matr{l^*}$.

\textbf{Graph Cut}\cite{boykov2004experimental} and the recently published \textbf{QPBO-I}\cite{rother2007optimizing} are fast exact solvers for the Maximum A Posteriori (MAP) of a binary MRF problem ($\mathcal{L} = \{0, 1\}$), as long as the energy terms have a \textbf{submodular} form, i.e. every pairwise term $\psi_{pq}$ satisfies
\begin{equation}
\psi_{pq}(0, 0) + \psi_{pq}(1, 1) \leq \psi_{pq}(0, 1) + \psi_{pq}(1, 0).
\end{equation}
\textbf{$\matr{\alpha\beta}$-swap} and \textbf{$\matr\alpha$-expansion}\cite{boykov2001fast} are iterative multi-label optimization schemes that \textit{approximate} problems of the form of equation \ref{eq:mrf} by iteratively minimizing binary MRFs with Graph Cuts or QPBO-I.

\begin{figure}[t]
\begin{center}
	\includegraphics[width=\linewidth]{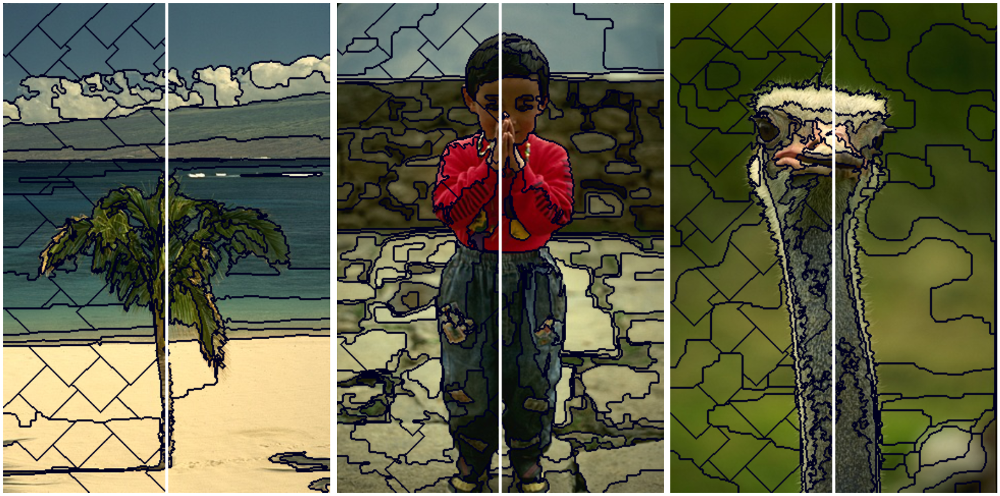}
\end{center}
   \caption{Sample result of our superpixel algorithm. The left part of each image is constrained to $200$ superpixels, the right side shows the same quality with much fewer superpixels.}
\label{fig:ours}
\end{figure}

\textbf{Anisotropic denoising} of a gray-scale image $\mathcal{I}$ can be formulated as a MRF-based pixel labelling problem by setting the set of labels $\mathcal{L}$ to $[0, 255]$ for all the possible gray values, the unary potentials $\psi_p(l_p) = (\mathcal{I}_p-l_p)^2$ to enforce the denoised image to be similar to the original image, the pairwise potentials $\psi_{pq}(l_p,l_q) = |l_p - l_q|$ to enforce smoothness between adjacent pixels by encouraging adjacent pixels to have similar labels, and the weight $w_{pq}$ inversely proportional to the gradient magnitude between $\mathcal{I}_p$ and $\mathcal{I}_q$ to avoid over-smoothing near the edges (ie. controls anisotropicity). Rewriting the MRF equation \ref{eq:mrf} for denoising yields
\begin{equation}\label{eq:denoise}
E(\matr{l}) = \sum_{p\in\mathcal{V}} (\mathcal{I}_p - l_p)^2 + \lambda\sum_{p,q\in\mathcal{E}} w_{pq}|l_p - l_q|.
\end{equation}
Here, the graph's nodes correspond to image pixels, edges correspond to 4-connected or 8-connected neighbours and $\lambda$ controls the strength of the denoising, as higher values of lambda produce higher denoising effects. The energy defined for the MRF-based denoising is submodular, and thus, can be efficiently solved with $\alpha\beta$-swap or $\alpha$-expansion methods.

\subsection{Piecewise-Constant Superpixels}\label{sec:superpixel}

The MRF formulation of color-image denoising can be easily extended from the gray-scale version. However, the computational cost of the fast approximate solvers quickly increases with the number of labels. MRF-based gray-scale denoising with $|\mathcal{L}|=256$ labels is already  computationally expensive; adapting it for a color image with $|\mathcal{L}|=256^3$ labels is in practice infeasible for a superpixel application.

It is known that a reduced set of a few quantized colors in the L*a*b space is sufficient to represent an image without confusing human visual perception. A more general application of feature quantization, namely the bag-of-words model, is currently widely applied in computer vision problems where a set of $K$ quantized features (with $K=100$ or $K=200$) over the whole dataset have enough discriminative power to create histogram features for classification and labelling.

Here we formulate the superpixel over-segmentation as a piece-wise constant denoising process in the quantized feature space. Given a color image $\mathcal{I}$ with $n$ pixels, its feature representation $\matr{X} = \{\matr{x}_1,\matr{x}_i,\dots,\matr{x}_n\}$ and a set of $K$ quantized features $\matr\Theta = \{\matr\theta_1,\matr\theta_i,\dots,\matr\theta_k\}$ from $\matr{X}$, the set of labels is defined as $\mathcal{L} = \{1,i,\dots,K\}$ and the unary and pairwise potentials are defined as
\begin{equation}\label{eq:unary}
\psi_p(l_p) = \norm{\matr{x}_p - \matr\theta_{l_p}}^2_2,
\end{equation}
\begin{equation}\label{eq:pairwise}
\psi_{pq}(l_p, l_q) = \norm{\matr\theta_{l_p} - \matr\theta_{l_q}}_1,
\end{equation}
\begin{equation}
w_{pq} = exp(-\gamma\norm{\matr{x}_p-\matr{x}_q}^2_2),
\end{equation}
where $l_p\in \mathcal{L}$ is the label of the pixel $p$ with $1\leq l_p\leq K$. Here, the unary potential $\psi_p$ enforces similarity between the pixel feature $x_p$ and the assigned quantized feature $\theta_{l_p}$ while pairwise potential $\psi_{pq}$ encourages similar adjacent pixels to have the same label. Our feature-denoising scheme can then be rewritten by introducing potentials from equations \ref{eq:unary} and \ref{eq:pairwise} in the general MRF formulation of equation \ref{eq:mrf} as
\begin{equation}\label{eq:fdenoise}
E(\matr{l}) = \sum_{p\in\mathcal{V}} \norm{\matr{x}_p - \matr\theta_{l_p}}^2_2 + \lambda\sum_{p,q\in\mathcal{E}} w_{pq}\norm{\matr\theta_{l_p} - \matr\theta_{l_q}}_1
\end{equation}
Note that for a gray-scale image $\mathcal{I}$, by setting $K=256$, $l\in\mathcal{L} = \{1,\dots,K\}$ for 256 gray-scale levels, $\matr{X} = \mathcal{I}$ and $\matr\Theta = \mathcal{L}$ we recover the original gray-scale denoising formulation from equation \ref{eq:denoise}.

To extract superpixels from color images, we set the feature vector $\matr{X}$ as the 3-feature color vector in the normalized\footnote{The image in L*a*b color space is rescaled to the [0,1] range.} L*a*b space where $\matr{x}_i = [L_i, a_i, b_i]$. To extract the $K$ quantized features, $M$ features are sampled from $\matr{X}$ without replacement and feed to a $k$-means algorithm that is initialized 10 times using the \textit{k-means++} \cite{arthur2007k} algorithm, from which results of the best initialization are taken (in terms of inertia). The $K$ cluster centers of the $k$-means clustering are chosen as the $\matr\Theta$ quantized features ($K$ quantized L*a*b colors). Following a bag-of-words study \cite{yang2007evaluating} we set $M = 10000$ random samples without replacement in our experiments, and $K$ is set to $16$ color features as 16 colors were sufficient in practice.

The energy from \ref{eq:fdenoise} is submodular and we approximate the globally optimal solution using $\alpha$-expansion with QPBO-I \cite{rother2007optimizing}.

\subsubsection{Extracting superpixels}\label{sec:extract}

The output of our feature denoising algorithm is a piece-wise constant graph, as neighbouring nodes are encouraged to have the same label $l\in \mathcal{L}$ (which corresponds to a quantized L*a*b color). By post-processing the result with a connected components algorithms that assigns neighbouring nodes with the same label to the same component, we are able to extract $N$ connected components which stand, in this case, for $N$ superpixels. This can be done very efficiently using a breadth-first search algorithm, which is similar to the post-processing applied in SLIC \cite{achanta2012slic} or LSC \cite{Li2015CVPR}, where small superpixels are also merged to their neighbours.

\subsubsection{Enforcing number of superpixels}

\begin{figure}[t]
\begin{center}
	\includegraphics[width=\linewidth]{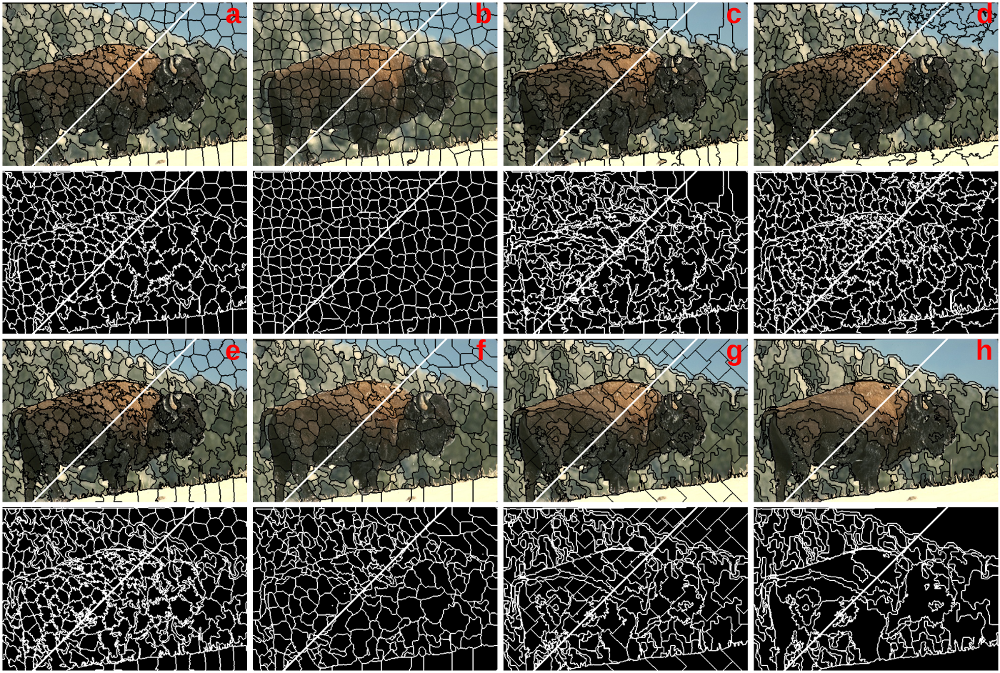}
\end{center}
   \caption{Comparison of segmentation algorithms with their corresponding boundary maps, with 400/200 superpixels. (a) SLIC, (b) Turbopixels, (c) SEEDS, (d) ERS, (e) LSC, (f) EneOpt1, (g) Ours (HPCS) and (h) Ours without enforcing the maximum size $s^\text{+}$ of superpixels (yielding fewer superpixels)}
\label{fig:comparison1}
\end{figure}

We enforce the number of superpixels in our solutions as a constrained post-processing step. For $S = W\times H$ the number of pixels in an image, and $N$ the number of desired superpixels, we define $s = S / N$ as the average size of the desired superpixels and $s^\text{-} = s{}/ 5$ and $s^\text{+} = s\cdot2$ as the minimum and maximum size of the desired superpixels respectively. The previous post-processing is then replaced with a two step split \& merge. 

The first step is a breadth-first search that finds connected components (connected sets of pixels with the same label) with a maximum size $s^\text{+}$ constraint. The search finishes the component when it reaches the maximum size and continues to the next one, splitting big segments in the process. The output of this one-pass step is a Region Adjacency Graph $\mathcal{R} = (\mathcal{V}_{sp}, \mathcal{E}_{sp})$ (RAG or graph of adjacent connected regions) where the nodes are $|\mathcal{V}| = T$ connected regions ($T$ superpixels) and edges connect neighbouring superpixels. Superpixels $p\in \mathcal{V}_{sp}$ are described by the constant label of its pixels $f_p\in\mathcal{L}$ and their size $s_{p}\in\mathcal{S}$ with $s_p\leq s^\text{+}$ (number of pixels that form it).

For the second step all nodes $p\in\mathcal{V}_{sp}$ with $s_{p} < s^\text{-}$ are merged to their most similar neighbour in the quantized label space. That is, a small node $p$ is merged with its neighbour $q$ such as
\begin{equation}
\argmin_{q\in\mathcal{N}(p)} \quad\psi_{pq}(f_p, f_q)
\end{equation}
where $f_p$ and $f_q$ stand for the label of the superpixel $p$ and its neighbour $q$ respectively. The merging of nodes $p$ and $q$ implies the following update steps,
\begin{equation}
\begin{split}
s_q = s_q + s_p,\\
\mathcal{V}_{sp} = \mathcal{V}_{sp} - \{p\},\\
\mathcal{E}_{sp} = \mathcal{E}_{sp} - \{(p, w)\mid \forall w\in\mathcal{N}(p), w \neq q\},\\
\mathcal{E}_{sp} = \mathcal{E}_{sp} \cup \{(q, w)\mid \forall w\in\mathcal{N}(p), w \neq q\},
\end{split}
\end{equation}
where $\mathcal{N}(p)$ refers to the neighbours of $p$. Node $p$ is removed from the graph and neighbours of $p$ are therefore relinked to $q$. The size of $q$ is updated accordingly with the size of $p$ and the merging step is iterated until no node is left with $s_p\leq s^\text{-}$. We obtain the optimal merge by ordering the edges $(p, q)\in\mathcal{E}_{sp}$ by their similarity $\psi_{pq}(f_p,f_q)$ and iteratively merging the most similar nodes $p$ and $q$ if either $s_p < s^\text{-}$ or $s_q < s^\text{-}$.

At the end of the two-step split \& merge post-processing, the RAG $\mathcal{R}$ maps each of the pixels from the image to a superpixel $sp\in V_{sp}$ where small nodes from $V_{sp}$ have been merged yielding $|V_{sp}| \simeq N$. Here $N$ stands for the desired number of superpixels.

As our algorithm doesn't rely on the number of superpixels as an important parameter initially, its computational cost is the same for any number of superpixels, as the post-processing step takes only around 1\% of the total computational time. By setting $s^\text{+} = S$ and $s^\text{-} = 0$ we recover the standard connected components post-processing from the section \ref{sec:extract} (where size and number of superpixel constraints are ignored) as none of the steps above neither split nor merge any superpixel. This simple post-processing allows us to generalize the superpixel algorithm to other applications where the number of superpixels is not relevant. It can be seen that while setting $s^\text{+} = s\cdot2$ provides state of the art compact results (results and discussion in section \ref{sec:experiments}), ignoring the maximum size constrain will give visually more appealing and non-compact results similar to the ones of the widely used mean-shift \cite{comaniciu2002mean} or Felzenwalb's efficient graph-based segmentation \cite{felzenszwalb2004efficient}. Thus, our algorithm can be easily set up to provide different kinds of superpixels, as required by the target application.

Figure \ref{fig:ours} shows an example results from our algorithm in the BSD500 dataset \cite{MartinFTM01}, while figure \ref{fig:comparison1} shows results of our superpixel algorithm in a sample image, with and without the maximum size constrain, compared to the top state of the art algorithms.

\subsection{Piecewise-Constant Supervoxels}

Supervoxel formulation is straight forward as all the steps of our algorithm are formulated as a graph-based framework. The only consideration need to made is whether to use 6/18/26-neighbour system in 3D instead of the standard 4/8-neighbour system in 2D images.

\subsection{Hierarchical super-region generalization}\label{sec:superregion}

Recently published work by L. Ladicky \etal \cite{russell2009associative} show that hierarchically generalized associative MRFs improve considerably the results in image semantic labelling by adding hierarchical contextual information. In their formulation, they create a MRF hierarchy by adding Higher Order potentials based on superpixels and what they term \textit{supersegments} (\textit{superpixels of superpixels}). They obtain superpixels by means of the Meanshift algorithm \cite{comaniciu2002mean}, and apply Meanshift again over the obtained superpixels to obtain \textit{supersegments}. However, it is known \cite{achanta2012slic} that the Meanshift algorithm for superpixel generation is quite slow, and recent state of the art algorithms can obtain considerably better results. Additionally, other superpixel algorithms can't be easily generalized hierarchically as they rely on the number of superpixels and some compactness constrain.

Here we generalize the superpixel and \textit{supersegments} terms into what we call ``super-regions''. Being super-regions hierarchically obtained from a ground set of pixels, we see the superpixels as a 1st-order super-regions, while the \textit{supersegments} from \cite{russell2009associative} as a 2nd-order super-regions. Here we will show that our algorithm can be generalized in a \textit{n}-th-order super-region framework.

For an \textit{i}-th-step in the hierarchy, the region adjacency graph from the previous iteration is taken as an input graph $\mathcal{G}^i$ and the input features $\matr{X}^i$ are set to be the mean color of each region from the $(i-1)$-th iteration's output, yielding
\begin{equation}\label{eq:hierar}
\mathcal{G}^i = \mathcal{R}^{i-1} \quad and \quad
\matr{X}^i = \matr{\mu}^{i-1}_{sp},
\end{equation}
and enforcing $K^i \leq K^{i-1}$. We have the particular case of the 0-th layer's output (corresponding to the input of the 1st layer, our superpixel algorithm in section \ref{sec:superpixel})
\begin{equation}
\mathcal{R}^0 = \mathcal{G}^0\quad and \quad \matr\mu^0_{sp} = \mathcal{I},
\end{equation}
where $\mathcal{I}$ is the input image and $\mathcal{G}^0$ is the graph of 4/8-neighbour system over the image $\mathcal{I}$.

It can be seen that the generalization is straight-forward, as the output of our algorithm (a RAG and the mean color of each super-region) serve as input for the next hierarchy level, where features are again quantized using \textit{k}-means (if needed), and the feature-denoising and post-processing steps can again be applied, yielding this time larger super-regions and another RAG $\mathcal{R}^i$, that can be used for further iterations.

The number of super-regions (or maximum and minimum size of the super-regions) and the $\lambda$ parameter that controls the denoising strength can be set up for every hierarchy level independently.

\begin{figure}[t]
\begin{center}
	\includegraphics[width=\linewidth]{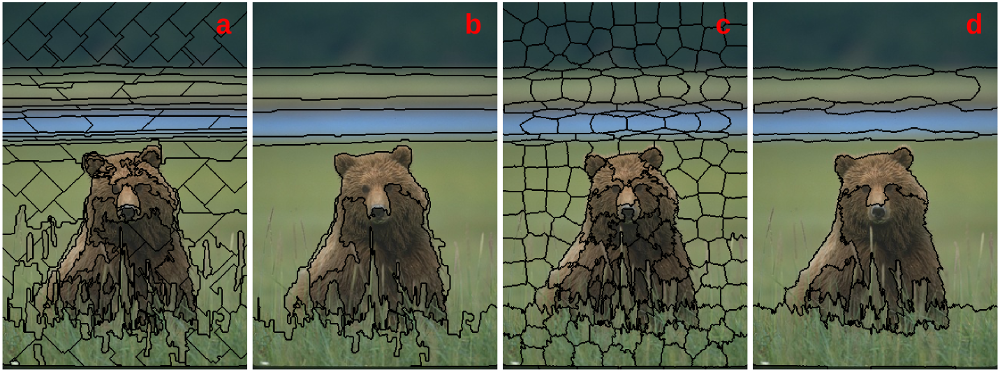}
\end{center}
   \caption{(a) our HPCS algorithm generating $200$ superpixels (1st-order super-regions), (b) our HPCS over-segmentation over (a) with 17 resulting 2nd-order super-regions, (c) SLIC superpixel over-segmentation with $200$ superpixels and (d) our HPCS over-segmentation over (c) resulting in 22 2nd-order super-regions. By applying our algorithm as a post-processing step the image can be represented with less superpixels.}
\label{fig:hirarchy}
\end{figure}

Figure \ref{fig:hirarchy} shows an example of our algorithm in a 2nd level hierarchy (computed twice recursively over the pixel grid) and an example of our algorithm as a post-processing step for reducing the number of superpixels (while keeping its properties) for other superpixel algorithms. In the example, the output of the SLIC superpixels is set as the input for our 2nd-order super-regions by extracting a RAG from SLIC superpixels $\mathcal{R}^{1}$, and assigning to each superpixel its mean color in the L*a*b as a feature $\matr\mu_{sp}$.


\section{Experiments}\label{sec:experiments}

We perform two experiments to evaluate qualitatively and quantitatively our super-region over-segmentation framework. In the first experiment, we compare our HPCS algorithm as a superpixel method to other state of the art superpixels algorithms in the BSD500 dataset \cite{MartinFTM01}. In the second experiment, we use our algorithm as a post-processing to reduce the amount of superpixels of results generated by the state of the art methods and we show that by doing so, we are able to represent an image with less superpixels, while potentially improving an algorithms' performance with lower amount of superpixels.

\subsection{Evaluation of HPCS as a superpixel algorithm}

\begin{figure*}
\begin{center}
	\includegraphics[width=0.95\textwidth]{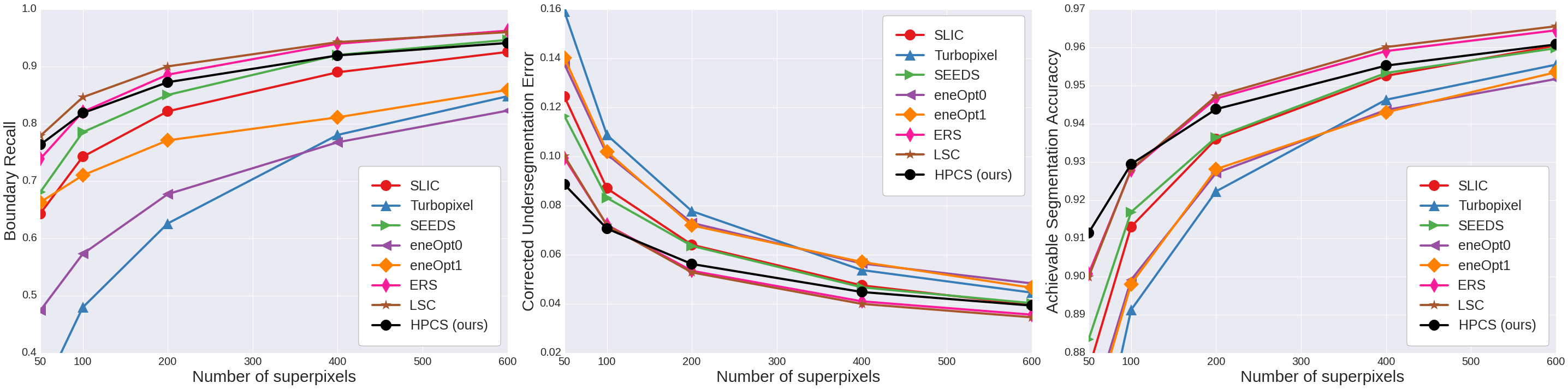}
\end{center}
   \caption{Quantitative evaluation our HPCS method as a superpixel algorithm.}
\label{fig:bsdbenchmark}
\end{figure*}

We compare 1st-order HPCS super-regions (that is, superpixels) to seven state-of-the-art superpixel algorithms: SLIC \cite{achanta2012slic}, Turbopixels \cite{levinshtein2009turbopixels}, SEEDS \cite{van2012seeds}, EneOpt0 and EneOpt1 \cite{veksler2010superpixels}, ERS \cite{liu2011entropy} and LSC \cite{Li2015CVPR}. For all the algorithms, we use the implementation publicly available in the author's web pages. We perform the experiments using the Berkeley Segmentation Dataset (BSD500) consisting of 500 images split into 200/100/200 training, validation and testing sets respectively, all with at least 4-5 manually segmented ground truth boundaries. We use the training set to empirically choose a default parameter for $\lambda$ (in all our experiments we use $\lambda = 0.1$), and we use the 200 images of the test set to calculate benchmarks and compare our algorithm to the others. We compare the quality of the superpixels by the commonly used three evaluation metrics: corrected under-segmentation error (\textbf{CUE}), boundary recall (\textbf{BR}) and achievable segmentation accuracy (\textbf{ASA}). Here, \textbf{CUE} measures the error of superpixels overlapping more than one ground truth object. Lower \textbf{CUE} indicates that fewer superpixels overlap more than one ground truth object. \textbf{ASA} measures the maximum achievable segmentation accuracy when using superpixels as units by assigning each superpixel to the object that it most overlaps. High values of \textbf{ASA} indicate that the over-segmentation matches well higher level objects. \textbf{BR} measures the fraction of ground truth boundaries that match superpixel boundaries. It is measured as the percentage of true boundary pixels that are within $2$ pixels from at least one superpixel boundary point. Here we adopt the definition of \textbf{CUE} used in \cite{van2012seeds} and \textbf{BR} and \textbf{ASA} from \cite{liu2011entropy}\cite{achanta2012slic}.

Figure \ref{fig:bsdbenchmark} shows the experimental results which are averaged over the 200 images in the test partition of the BSD500 dataset. Despite its simple formulation, it can be seen that our algorithm is in general as good or better than most state of the art algorithms, especially with lower numbers of superpixels. It is however, slightly more slower that some of the algorithms. Our algorithm takes 2-3 seconds using a standard i3 desktop computer for each image in the BSD500 dataset, which is slower than SEEDS, SLIC and LSC that take less than a second, but is still as fast as ERS (which takes around 3 seconds) and faster than Turbopixels and eneOpt0-1 (on average \textgreater 5 seconds). Despite  eneOpt0-1 being also formulated in a MRF framework, its gray-scale formulation (to make it efficient) makes it worse in terms of boundary adherence. This is probably due to the loss of the discriminative information that the color provides. Our algorithm, as with LSC and ERS, obtains superpixels in a global formulation as an approximation to the global optimal solution from equation \ref{eq:fdenoise} and our post-processing step. This can be seen in the benchmark as they obtain sightly better results than SEEDS and SLIC, which rely on local features, although the results are close.

\begin{figure*}
\begin{center}
	\includegraphics[width=0.95\textwidth]{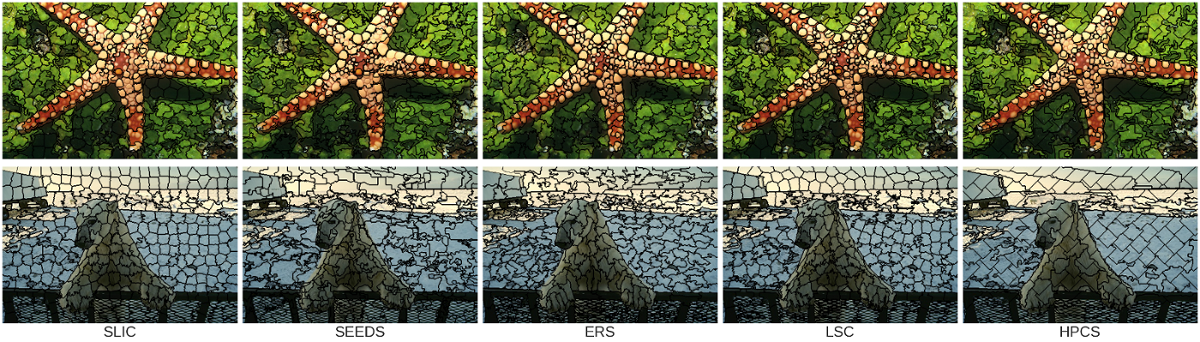}
\end{center}
	\caption{Visual comparison of superpixels over-segmentation results for $N = 400$ superpixels.}
\label{fig:overview}
\end{figure*}

In all the above benchmarks, as different runs of our algorithm might obtain sightly different results due to the initial \textit{k}-means feature quantization, we ran our algorithm 5 times per image and averaged the scores. We found, however, that as for each image we quantize the features by running 10 \textit{k}-means with \textit{k-means++} and keep the best solution, the difference between initializations is minimal and all of the end results preserve the same strong object boundaries. Figure \ref{fig:overview} shows qualitative comparison of 5 selected superpixel algorithms for $N = 400$ superpixels.

\subsection{Evaluation of hierachical HPCS applications}

To evaluate our hierarchical super-region formulation, we obtain 2nd-order super-regions and calculate the same evaluation metrics \textbf{BR}, \textbf{CUE} and \textbf{ASA} as in the previous section. We empirically show that, by applying our algorithm over a previously obtained superpixel over-segmentation, we are able to reduce the number of superpixels while maintaining higher level objects. It is difficult to evaluate the high level generalization properties of our algorithm, as ground truth boundaries from BSD500 come from different users and vary a lot in the higher level detail of the delineated objects (i.e. some manual ground truth boundaries would contain a whole \textit{car} as an object, while other ground truth boundaries split a \textit{car} in several parts). Thus, we empirically demonstrate the effectiveness of our method by obtaining SLIC superpixel over-segmentations with $N=400$ superpixels and use them as an input to apply our 2nd-order HPCS over them to reduce the amount of superpixels to an average of $26$. We show that, by doing this, we obtain $26$ super-regions that obtain same or better results than the original algorithm with $N=50$ superpixels.

\begin{table}
\begin{tabular}{c|cccc}
 & \textbf{Num SP} & \textbf{BR} & \textbf{CUE} & \textbf{ASA}\\
\hline
SLIC50          & 40 & 0.643 & 0.124 & 0.876 \\
SLIC400 + HPCS  & \textbf{26} & \textbf{0.650} & 0.126 & 0.875 \\
\hline
\end{tabular}
\caption{Quantitative evaluation of 2nd-layer oversegmentation.}
\label{tbl:quant}
\end{table}

\begin{figure*}
\begin{center}
	\includegraphics[width=0.95\textwidth]{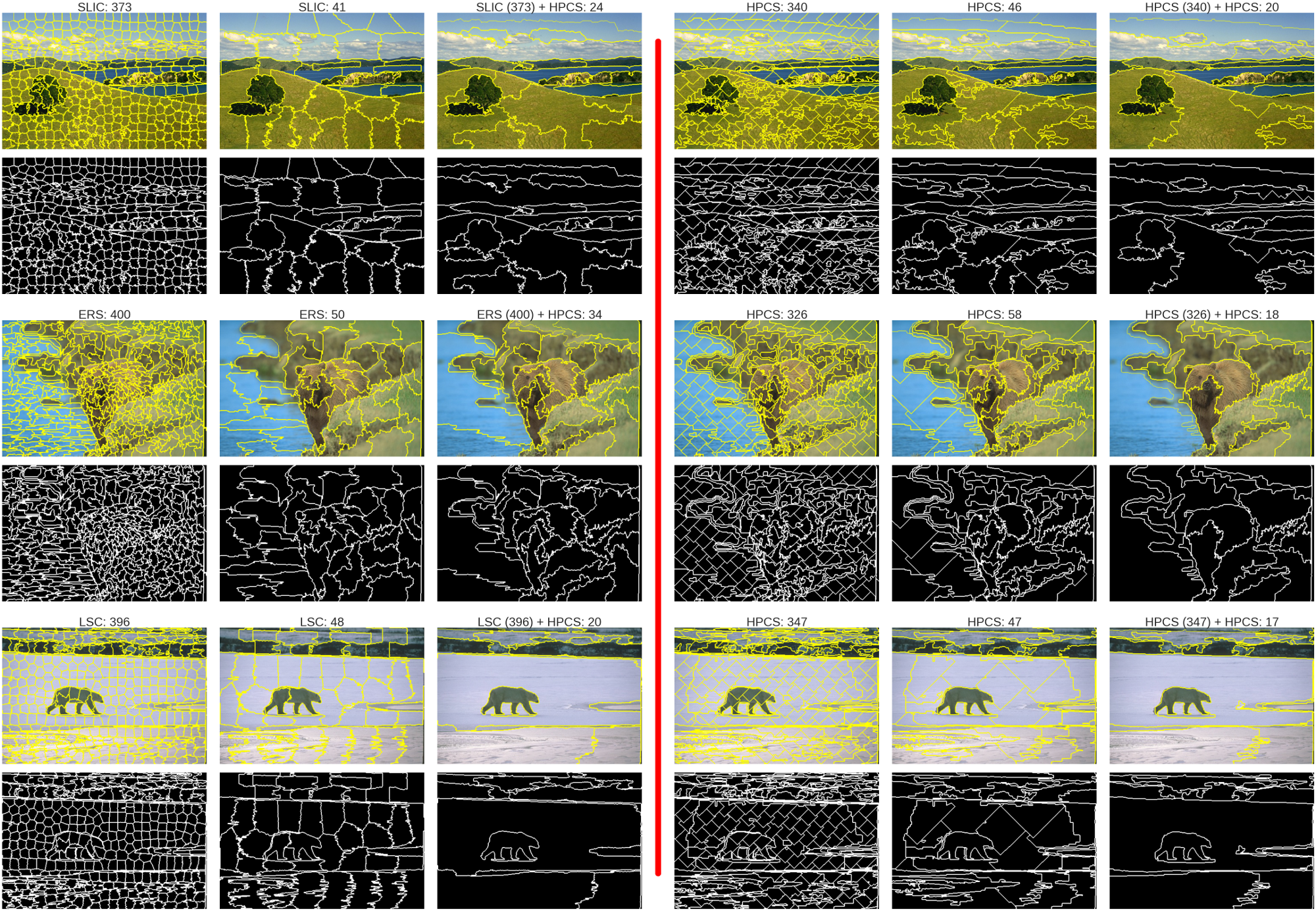}
\end{center}
	\caption{Qualitative comparison of over-segmentations in 3 different images from the BSD500 dataset. Each image is shown in 2 consequtive rows (the first one contains the superpixel result, while the next its corresponding boundary map). In the left side, for each image a random algorithm from the literature is chosen. The first 2 columns show segmentation results with the selected algorithm for 400 and 50 superpixels respectively. The 3rd column shows $\sim 30$ 2nd-order super-regions by using the selected algorithm (with 400 superpixels) as an input for the second-layer of our HPCS algorithm. Right side of the figure shows the same procedure by applying twice recursively our HPCS super-regions. Numbers indicate the exact amount of super-regions in the image.}
\label{fig:hcomp}
\end{figure*}

Table \ref{tbl:quant} shows quantitative results by applying the above procedure over the 200 images of the BSD500 test dataset and obtaining the mean \textbf{BR}, \textbf{CUE} and \textbf{ASA} while figure \ref{fig:hcomp} shows qualitative results of our algorithm as a post-processing step for state of the art algorithms (including ours recursively). This second layer step is much faster than the superpixel generation, as the optimization is made over the superpixel graph, and thus, takes less than half a second per image.

\section{Conclusions}

In this paper we have presented a new superpixel formulation in an energy minimization framework that can be applied both hierarchically to obtain higher-level segmentations, and as a post-processing step for other superpixel methods. Our algorithm tends to form big superpixels where the is no characterizing texture (like ground or sky areas), and will create more superpixels in areas that require more attention to detail. We believe this has some interesting implications, as further layers in the hierarchy can delineate higher level objects. This however, requires further study and we plan to examine its application to hierarchical semantic segmentation \cite{russell2009associative} and in high dimensional image segmentation. Furthermore, for fast superpixel/super-region generation we only use color and mean color features; however, as the algorithm is formulated in a quantized feature space, any potential dense feature (such as hitograms, SIFT or filter banks) could be used with minor modifications. Experimental results validate our framework, both quantitatively and qualitatively.


{\small
\bibliographystyle{ieee}
\bibliography{mybib}
}

\end{document}